\documentclass[]{fairmeta}

\usepackage{microtype}
\usepackage{booktabs} 

\usepackage{hyperref}
\usepackage[utf8]{inputenc} 
\usepackage[T1]{fontenc}    

\usepackage{url}            
\usepackage{amsfonts}       
\usepackage{nicefrac}       
\usepackage{xcolor}         
\usepackage{bm}
\usepackage{bbm}
\usepackage{amsthm}
\usepackage{psfrag,thmtools}
\usepackage{amssymb}
\usepackage{mathrsfs}
\usepackage{multirow}
\usepackage{xspace}
\usepackage{dsfont}
\usepackage{natbib}
\usepackage{wrapfig}
\usepackage{enumitem}
\usepackage{booktabs}
\usepackage{yfonts}
\usepackage[euler]{textgreek}
\usepackage[normalem]{ulem}
\usepackage[most]{tcolorbox}

\usepackage[algo2e]{algorithm2e}

\usepackage{amsmath}
\usepackage{mathtools}
\setlist{leftmargin=*, labelsep=5pt}
\theoremstyle{plain}
\newtheorem{theorem}{Theorem}[section]

\theoremstyle{definition}
\newtheorem{definition}[theorem]{Definition}

\theoremstyle{remark}

\def\MBP{{MBD}}

\title{{\MBP}: A Model-Based Debiasing Framework Across User, Content, and Model Dimensions}

\author[1,\dagger]{Yuantong Li}
\author[1,\dagger]{Lei Yuan}
\author[1]{Zhihao Zheng}
\author[1]{Weimiao Wu}
\author[1]{Songbin Liu}
\author[1]{Jeong Min Lee}
\author[1]{Ali Selman Aydin}
\author[1]{Shaofeng Deng}
\author[1]{Junbo Chen}
\author[1]{Xinyi Zhang}
\author[1]{Hongjing Xia}
\author[1]{Sam Fieldman}
\author[1]{Matthew Kosko}
\author[1]{Wei Fu}
\author[1]{Du Zhang}
\author[1]{Peiyu Yang}
\author[1]{Albert Jin Chung}
\author[1]{Xianlei Qiu}
\author[1]{Miao Yu}
\author[1]{Zhongwei Teng}
\author[1]{Hao Chen}
\author[1]{Sunny Baek}
\author[1]{Hui Tang}
\author[1]{Yang Lv}
\author[1]{Renze Wang}
\author[1]{Qifan Wang}
\author[1]{Zhan Li}
\author[1]{Tiantian Xu}
\author[1]{Peng Wu}
\author[1]{Ji Liu}

\affiliation[1]{Meta AI} 

\contribution[\dagger]{Equal Contribution}

\abstract{
Modern recommendation systems rank candidates by aggregating multiple behavioral signals through a value model. 
However, many commonly used signals are inherently affected by heterogeneous biases.
For example, watch time naturally favors long-form content, loop rate favors short-form content, and comment probability favors videos over images.
Such biases introduce two critical issues: (1) value model scores may be systematically misaligned with users' relative preferences — for instance, a seemingly low absolute like probability may represent exceptionally strong interest for a user who rarely engages; and (2) changes in value modeling rules can trigger abrupt and undesirable ecosystem shifts. 
In this work, we ask a fundamental question: can biased behavioral signals be systematically transformed into unbiased signals, under a user-defined notion of ``unbiasedness'', that are both personalized and adaptive?
We propose a general, model-based debiasing ({\MBP}) framework that addresses this challenge by augmenting it with distributional modeling. 
By conditioning on a flexible subset of features (partial feature set), we explicitly estimate the contextual mean and variance of the engagement distribution for arbitrary cohorts (e.g., specific video lengths or user regions) directly alongside the main prediction. 
This integration allows the framework to convert biased raw signals into unbiased representations, enabling the construction of higher-level, calibrated signals (such as percentiles or z-scores) suitable for the value model. Importantly, the definition of unbiasedness is flexible and controllable, allowing the system to adapt to different personalization objectives and modeling preferences. Crucially, this is implemented as a lightweight, built-in branch of the existing multi-task multi-label (MTML) ranking model, requiring no separate serving infrastructure.
We demonstrate the effectiveness of this framework through large-scale deployment on two billion-user-level short-video applications. Empirical results from rigorous online A/B testing show that {\MBP} significantly improves long-term retention and engagement metrics (e.g., Time Spent, Sessions) by successfully decoupling preference signals from intrinsic ecosystem biases.
}

\date{\today}
\correspondence{\email{yuantongli@meta.com}}

\begin{document}

\maketitle

\section{Introduction}
\label{section:intro}
Recommendation systems have become fundamental components of modern digital platforms, particularly with the exponential growth of short-form video content on platforms such as TikTok, Instagram Reels, Facebook Reels, and YouTube Shorts, managing content distribution to billions of daily active users \citep{davidson2010youtube, wang2017deep, liu2019user, wang2021dcn, quan2023alleviating, chen2023bias, jing2024multimodal, yang2025swat}. 
Modern recommendation systems serve users by aggregating a diverse collection of behavioral signals (e.g., likes, shares, watch time). However, these raw signals are intrinsically confounded by heterogeneous biases \citep{schnabel2016recommendations,wei2021model,chen2023bias, gao2024causal}. Specifically, these biases arise from three primary sources: (1) Item bias, where physical properties like video duration mechanically inflate engagement metrics regardless of content quality; (2) User bias, where some users are naturally more active or patient than others, and demographic factors like location and language often drive users to follow trends, causing popularity bias; and (3) Model bias, where the ranking system itself exacerbates these initial biases through feedback loops, progressively narrowing the ecosystem to favor system-selected winners over genuine user interests.

To mitigate these issues, the industry has adopted various debiasing strategies, ranging from statistical approaches like inverse propensity weighting and statistical bucketing to advanced causal inference methods \citep{wang2021deconfounded,chen2023unbiased}. 
Recent advances in watch time prediction have introduced sophisticated debiasing techniques. Weighted logistic regression  \citep{covington2016deep} pioneered duration-aware prediction for YouTube. D2Q \citep{zhan2022deconfounding} mitigates duration bias through backdoor adjustments and quantile modeling under different duration groups. D2Co \citep{zhao2023uncovering} addresses both duration bias and noisy watching behavior, while DVR \citep{zheng2022dvr} introduces the watch time gain metric with adversarial learning for unbiased preference estimation. Despite their effectiveness in specific scenarios, all these methods lack a unified mechanism to estimate the contextual baseline distribution.

While effective in specific scenarios, these methods share a fundamental limitation: they operate primarily within a point-wise estimation paradigm. Standard ranking models aim to estimate the absolute expectation of engagement (e.g., ``this video will get 45 seconds of watch time''). However, a single point estimate lacks the contextual information necessary for fair comparison. For instance, 45 seconds of watch time indicates high interest for a 60-second clip but low interest for a 10-minute video. Traditional methods fail to quantify the contextual distribution of the baseline behavior. Without modeling the underlying uncertainty within a specific context (e.g., ``what is the normal watch time for this user on videos of this length?''), point-wise predictions cannot distinguish between genuine user interest and bias-driven inflation.

In this work, we propose model-based debiasing ({\MBP}), a generalized framework that addresses these limitations by augmenting the traditional point-wise ranking paradigm with distribution-wise characterization.
The core intuition of {\MBP} is to explicitly model the statistical properties of ``normal'' behavior for any given context. By defining a flexible partial feature set (e.g., \{user profile, video length\}), the framework simultaneously estimates the contextual mean ($\mu$) and variance ($\sigma^2$) of the engagement distribution directly within the ranking architecture. This allows us to systematically transform biased, absolute predictions into calibrated, relative signals (such as z-scores or percentiles). For example, instead of predicting a raw ``45 seconds'', {\MBP} allows the system to interpret the signal as ``the 85th percentile of performance for this specific video length'' (as detailed in Section \ref{sec:application}). Crucially, {\MBP} is implemented as a lightweight, built-in branch of the existing multi-task multi-label model, estimating these statistics via a dual-prediction framework without requiring separate serving infrastructure or offline statistical tables.

The contributions of this work are summarized as follows:

\begin{itemize}
	\item \textbf{Generalized Debiasing Framework}: We introduce a theoretical framework that shifts from point-wise error minimization to distributional bias mitigation. By utilizing partial feature sets, {\MBP} provides a unified solution for diverse bias types, including video length, user activity skew, and content cold-start.
	
	\item \textbf{Distribution-Free Learning Algorithm}: We propose the decoupled method of moment learning algorithm to estimate distributional statistics, offering distribution-free property.
	
	\item \textbf{Efficient Built-in Architecture}: We design a dual-prediction model architecture that integrates distribution modeling as an additional task sharing part of the main model's feature representation. This design incurs negligible engineering overhead and allows for real-time, context-aware debiasing.
	
	\item \textbf{Industrial Scale Impact}: We demonstrate the effectiveness of {\MBP} through large-scale deployment on a platform serving billions of users. Online A/B testing confirms significant gains in long-term engagement, including a cumulative increase of over +0.5\% in Time Spent and +0.05\% in Sessions across different product surfaces, validating that debiased signals effectively drive sustainable ecosystem growth.
\end{itemize}

\section{Preliminary: Existing Systems}
\noindent 
\begin{figure*}
	\centering
	\includegraphics[width=0.9\linewidth]{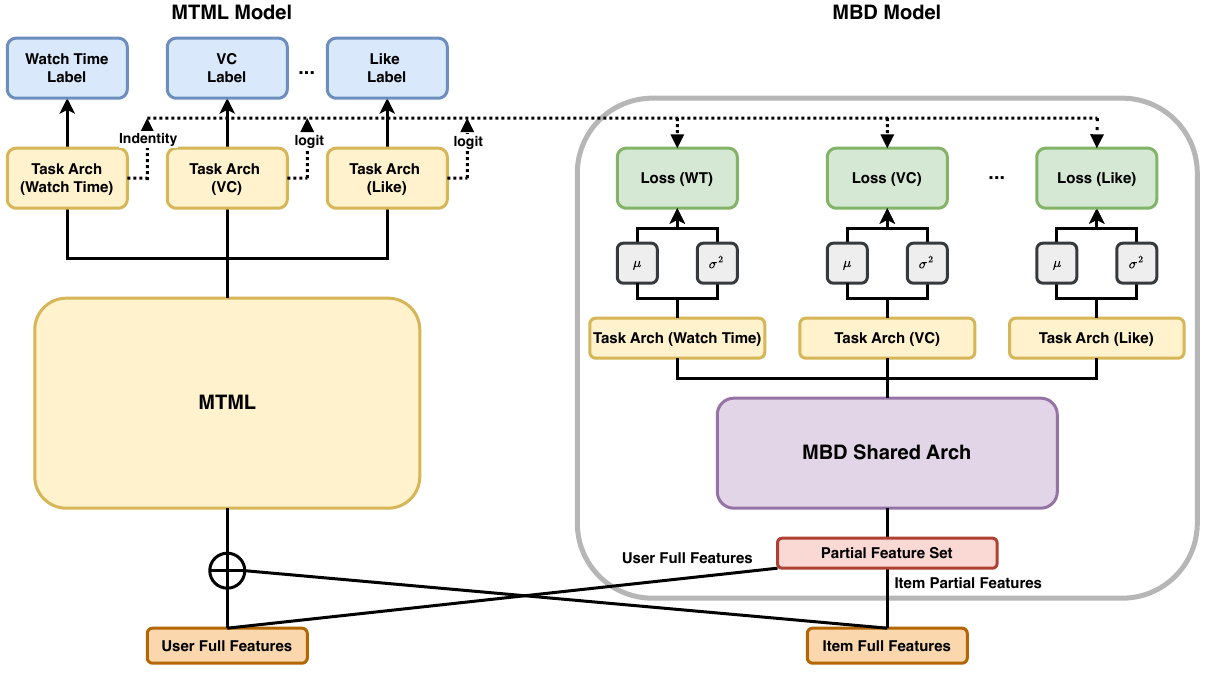}
	\caption{The {\MBP} framework augments the MTML model with an additional branch that estimates contextual distribution statistics $(\mu, \sigma^2)$ through a partial feature set.}
	\label{fig:mbp_arch}
\end{figure*}

This section introduces the typical recommendation system and how it may handle biased signals.

Let $\mathcal{U}$ denote the user set and $\mathcal{V}$ denote the item set. When there is an interaction between user $u$ and item $v$, the platform observes a ground-truth label $y_{u,v}$. Depending on the specific ranking task, this target variable may be binary (e.g., $y \in \{0, 1\}$ for likes or skips) or continuous (e.g., $y \in \mathbb{R}^+$ for watch time).
We define a feature mapping function $g(u, v)$ that extracts a dense feature representation $\mathbf{h} \in \mathbb{R}^d$, based on the user profile attributes, item content features (e.g., video duration, topic), and historical interactions. Additionally, let $\mathbf{x} = \{u,v\}$ denote the full features of the user and item. 

\subsection{Signal bias widely exist in typical recommendation systems}
The standard approach employs a multi-task multi-label (MTML) ranking architecture. The primary objective of this formulation is to accurately estimate contextual expectation of a target variable $Y_{u,v}$ given features $\mathbf{x}$:

\begin{equation}
	\label{eq:point-est}
	p(\mathbf{x}) = \mathbb{E}[Y_{u,v} \mid \mathbf{x}]
\end{equation}

For binary tasks, this corresponds to the probability of engagement $P(y=1|\mathbf{x})$, while for regression tasks, it represents the expected magnitude (e.g., predicted watch time in seconds).

\paragraph{Value Model (VM).} To generate the final recommendation list, the system merges $T$ tasks' predictions into one final score. This is achieved through a VM, which aggregates the heterogeneous vector of point-wise predictions $\vec{p}(\mathbf{x}) = \hat{\mathbf{y}} = [\hat{y}_1, \dots, \hat{y}_T]^\top$ into one score $S_{\text{final}}$.
The most prevalent industrial formulation is a linear weighted combination of transformed signals $S_{\text{final}} = \sum_{t=1}^{T} w_t \cdot \hat{y}_t$, 
where $w_t$ represents the hyperparameter weight for the $t$-th task, determined by business objectives. 

\paragraph{Inherent Biases in Signal Aggregation.}
While point-wise estimates successfully minimize prediction error, the subsequent linear aggregation introduces significant systematic bias by treating the raw engagement $\hat{y}_t$ as a direct proxy for user preference intensity. In reality, these signals possess heterogeneous functionality in business and are inherently confounded by content attributes rather than solely reflecting user interest. For instance, $\hat{y}_{\text{watch time}}$ is naturally higher for longer videos regardless of actual satisfaction, while loop rates favor shorter content. Consequently, the linear sum $S_{\text{final}}$ inadvertently amplifies intrinsic biases, aligning the final ranking with ecosystem artifacts rather than the user's true relative preference efficiency. Here we list several examples of bias:
\begin{itemize}
	\item Format Bias (video vs. photo): Watch time naturally favors video content over static photos regardless of user satisfaction. A 5s view on a photo might indicate high interest, whereas on a video it may be not interested, which is very common in feed scenario.
	\item Duration Bias (loop rate): The probability of looping ($p_{\text{loop}}$) naturally favors short videos over long videos. Directly comparing raw $p_{\text{loop}}$ scores unfairly suppresses longer, high-quality videos.
	\item User Threshold Bias: Different users possess different baseline tendencies for engagement (e.g., ``fast scrollers'' vs. ``patient watchers''), making absolute predictions incomparable across users.
\end{itemize}

\subsection{Naive Debiasing Approaches via Bucketized Counting}
Recognizing the systemic risks and impacts that biased signals introduce into recommendation systems, a natural and straightforward approach is to apply bucketized counting, although this approach has not been explicitly discussed in prior literature, to the best of our knowledge. This method aims to construct a relative signal by normalizing predictions against discrete group averages derived from historical data. Taking watch time duration bias as an example, the process involves three steps:

1. \textbf{Bucketing}: Items are grouped into discrete buckets based on the bias attribute (e.g., video duration intervals: 0-5s, 5-10s, 10-15s).

2. \textbf{Counting}: Compute the average historical engagement mean $\mu_k$ and variance $\sigma_{k}$ for all items falling into bucket $k$.

3. \textbf{Correction}: The original point-wise prediction is adjusted by mean or mean-variance of this group's baseline: 
\begin{align}
\hat{y}_{\text{unbiased}} = \hat{y}(u, v) - \mu_k, \text{or } \frac{\hat{y}(u, v) - \mu_k}{\sigma_{k}} 
\label{eq:naive}
\end{align}

\begin{figure}[ht]
    \centering
    \includegraphics[width=0.7\linewidth]{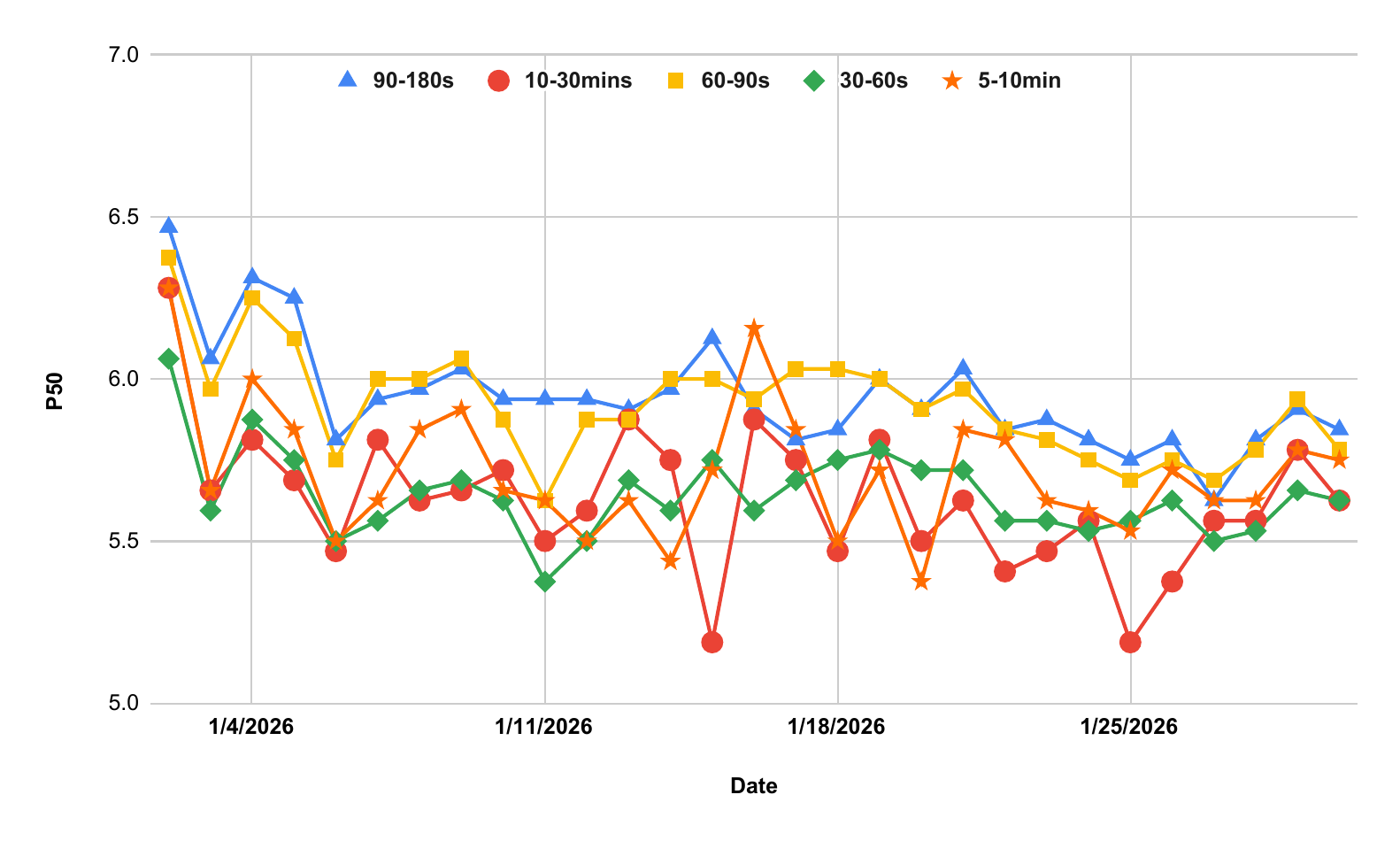}
    \caption{Transformed watch time ($P_{50}$) trend over one month bucketed by video length.}
    \label{fig:p50-trend}
\end{figure}

\paragraph{Limitations.} While statistical bucketing method provides a form of debiasing, it suffers from severe structural limitations when applied to large-scale personalized recommendation system. We categorize these failures into four dimensions:
\begin{itemize}
    \item \textbf{Discretization Error and Intra-Bucket Bias:} Bucketing approximates continuous bias curves with discrete step functions. This approach fails to capture bias within a specific bucket. For example, in a ``5s-10s'' duration bucket, a 5.1s video and a 9.9s video are treated as having the same baseline, despite the 9.9s video naturally accruing significantly higher watch time. This residual bias continues to distort ranking within the cohort.

    \item \textbf{The Curse of Dimensionality (Scalability):} Statistical bucketing methods rely on pre-computed statistical tables, which do not scale to high-dimensional contexts. To debias a signal based on multiple confounders simultaneously—such as $\text{user region} \times \text{video type} \times \text{duration}$ — the system would need to maintain a combinatorial explosion of buckets. This becomes computationally prohibitive and impossible in real scenario.
    
    \item \textbf{Data Sparsity and Generalization:} Statistical methods require sufficient historical data to establish a stable baseline ($\mu_k$). Consequently, they fail for cold start scenarios (infrequent users or fresh content) where the bucket is empty or sparse. It cannot construct a reliable signal for rare or unseen cross-combinations.

    \item \textbf{Temporal Staleness and Infrastructure Overhead:} Finally, heuristic baselines suffer from a critical lag. The baseline statistic $\mu_k$ is determined by a historical snapshot — typically aggregated from offline logs. This approach relies on the assumption that the engagement distribution is stationary. However, live ecosystems exhibit distributional drift due to trending topics, time-of-day effects. As illustrated in Figure \ref{fig:p50-trend}, global median watch time demonstrates a non-trivial downtrend over a period. Consequently, a baseline $\mu_k$ calculated from a ``past snapshot'' becomes systematically misaligned with the current inference - time distribution, leading to persistent bias (e.g., over-demoting content when the global average naturally drops).
    While real-time streaming aggregation could theoretically mitigate this, it incurs prohibitive engineering complexity and synchronization latency.
\end{itemize}

\section{{\MBP}: Model-Based Debiasing Framework}
To systematically address these structural deficiencies, we propose {\MBP}, a generalized framework designed to overcome the limitations of the typical recommendation system.

Unlike bucketized counting — which is plagued by the curse of dimensionality, intra-bucket bias, data sparsity, and temporal staleness — {\MBP} provides a unified solution by creating a partial feature set to model. 
This allows the system designer to specify arbitrary bias dimensions to debias existing signals. This framework includes three key components: 1) Specify the bias feature set - a subset of features (we want to remove bias from them, e.g., video length, user nation); 2) Learning debiased mean and variance over the specified bias set admitting maximal efficiency and reliability by adding a cheap computational module onto the regular ranking model but with zero impact to ranking model; 3) Construct the unbiased signals from bias mean and bias variance together original biased signal.


\subsection{Specifying Bias}

\begin{definition}[\textbf{Bias Feature Set}]
    Let $\mathbf{x}$ denote the full feature set encompassing all user and item attributes. We define a partial feature set $\mathbf{x}'$ as an arbitrary subset representing the specific bias factors we wish to control for.
\end{definition}

This formulation offers great flexibility. We can choose arbitrary set of features as the debiasing target. For example, $\mathbf{x}'$ can be configured to represent:
\begin{itemize}
    \item Duration Debiasing: $\mathbf{x}' = \{ \mathbf{u}_{\text{full}}, v_{\text{length}} \}$. Here, the baseline is conditioned on the user's full profile and the video length, allowing comparisons relative to the user's typical watch time for that duration.
    \item Content Cold-Start Debiasing: $\mathbf{x}' = \{ \mathbf{u}_{\text{full}}, v_{\text{views}} \}$. The baseline is conditioned on the item's exposure level (e.g., the number of views on new videos), enabling fairness adjustments for new content lacking historical interaction data.
    \item Regional Debiasing: $\mathbf{x}' = \{u_{\text{region}}, \mathbf{v}_{\text{full}} \}$. The baseline is conditioned on the user's geographic cohort (e.g., US/CA vs. ROW) and item attributes, accounting for regional variations in engagement volatility.
\end{itemize}

\subsection{Contextual Mean and Variance Estimates on Bias Set via Supervised Learning}
\label{sec:mbp-estimation}

The essential problem is to estimate contextual mean and variance over the bias feature set $\mathbf{x}'$ for specific task,
\begin{equation}
    \begin{aligned}
        \mathbb{E}[y_{u,v} \mid \mathbf{x}'] \\
        \text{Var}[y_{u,v} \mid \mathbf{x}']
    \end{aligned}
\end{equation}

The mean estimate can be easily formulated into a supervised learning problem  like Eq.\eqref{eq:point-est}. 
\begin{align}
\label{eq:mean-est}
\min_{\mu} \mathcal{L}_{\text{mean}}(\mu) = [\text{sg}[p(\mathbf{x})] - \mu(\mathbf{x}')]^2 
\end{align}
and the solution is $\mathbb{E}[\text{sg}[p(\mathbf{x})] \mid \mathbf{x}']$.

In comparison, the estimate of contextual variance is not straightforward because the supervised signals do not exist unlike the contextual mean prediction. To make contextual variance learnable under the supervised learning framework, we are inspired by the relation among the variance,  the 1st order moment, and the 2nd order moment
\begin{align}
\label{eq:var-est}
\text{Var}(X) = \mathbb{E}[(X -\mathbb{E}[X])^2]
\end{align}
where $X$ is a random variable. Then we formulate this contextual variance estimate problem into a supervised learning problem by the following loss function
\begin{equation}
\begin{aligned}
    \min_{\sigma^2} \mathcal{L}_{\text{var}}(\sigma^2) &= \| \sigma^2(\mathbf{x}') - \text{sg}\left[ [p(\mathbf{x}) - \text{sg}[\mu(\mathbf{x}')]]^2 \right] \|^2
\end{aligned}
\label{eq:variance est}
\end{equation}

where $\text{sg}[\cdot]$ denotes the stop-gradient operator.

It is worth noting that 1) this approach is consistent with the bucketized counting approach when the bucket is granular, so {\MBP} can be considered to be a generation of the naive bucketized counting approach; 2) this estimate framework does not assume any underlying distribution. 

\textbf{Efficiency and Reliability.} These two supervised loss terms for contextual mean and variance estimates can be added to the main ranking loss as the additional ones, maximally reusing the intermediate embedding of the main ranking model if we can share user or item embedddings. Thus, the overall increased computational cost is typically below 5\%. To avoid the unanticipated impact to the main ranking model, the components for contextual mean and variance estimates only update their own parameters by detaching the gradient backpropagation from the original module in the ranking model. 

\textbf{Debiasing Ranking Model.} The contextual mean and variance signals are going to be used together with the original predictive signal, for example, in the way in Eq. \eqref{eq:naive}. In the real scenario, the original predictive signal could be significantly deviated from the contextual mean and variance due to unanticipated stability issue or calibration issue. To avoid them, a more robust choice of labels in Eq.\eqref{eq:mean-est} and Eq.\eqref{eq:variance est} are the prediction by the ranking model rather than the ground truth labels.

The overall model architecture can be found in Figure~\ref{fig:mbp_arch}.

\subsection{Unbiased Signal Construction from Contextual Mean and Variance}
The contextual mean and variance estimates on the predefined bias set $\mathbf{x}'$ allow us to construct unbiased signals together the original (biased) signal. Per application scenarios, there are various approaches to construct the unbiased counterpart from the biased signal. In the following, we provide a few examples. All of the admits a common principle - removing the bias components (defined by contextual mean and variance) from the original bias signal. The most common construction is the relative preference score (RPS) similar to Eq. \eqref{eq:naive}, defined as:
\begin{equation}
    \text{RPS} = \frac{p(\mathbf{x}) - \mu(\mathbf{x}')}{\sigma(\mathbf{x}')}
\end{equation}
This transformation effectively converts a raw prediction (e.g., ``45 seconds'') into a standardized metric (e.g., ``85th percentile performance'' if compound with Gaussian CDF $\Phi(\cdot)$), enabling fair comparison across heterogeneous content types.


Here we provide three examples to illustrate how to integrate the RPS into the VM system.

\textbf{1. Additive Boosting (High-Confidence Promotion):}
This strategy is designed to reward items that perform above their cohort baseline. We adopt a hinge-based boosting formulation to ensure only boosting candidates that exhibit ``high confidence'' while leaving average items untouched.
\begin{align}
    S_{\text{final}} = S_{\text{final}} + w \cdot \max(0, \text{RPS} - \tau_{\text{high}})
\end{align}
where $\tau_{\text{high}}$ is a statistical significance threshold (typically $\tau = \mu + \alpha \sigma$, with $\alpha \in [1.0, 3.0]$). This effectively targets top-tier content for promotion without introducing noise from the dense central distribution.

\textbf{2. Hard Filtering (Low-Confidence Suppression):}
To safeguard user experience against low-quality content (e.g., clickbait), we employ an indicator-based gating mechanism. This strategy suppresses candidates that fall significantly below their cohort expectations.
\begin{align}
    S_{\text{final}} = S_{\text{final}} \cdot \mathbb{I}(\text{RPS} \geq \tau_{\text{low}})
\end{align}
where $\mathbb{I}(\cdot)$ is the indicator function and $\tau_{\text{low}}$ represents a minimum quality floor (e.g., $\tau = \mu - \beta \sigma$).

\textbf{3. Multiplicative Reweighting (Soft Calibration):}
For scenarios requiring continuous calibration rather than binary thresholds, we utilize a ratio-based or sigmoidal approach. This acts as a ``soft reweighting'' mechanism that adjusts the score magnitude relative to the item's baseline performance.
\begin{align}
    S_{\text{final}} = S_{\text{final}} \cdot q_{\text{scale}}(\text{RPS})
\end{align}
where $q_{\text{scale}}$ is typically a normalized scaling function (e.g., $(p / \mu)^\alpha$ or $\text{Sigmoid}(z)$). This strategy preserves the relative order of candidates but expands or compresses the score distribution to reflect the true relative preference efficiency across different cohorts.

\section{Experiments}
\label{sec: experiments}

In this section, we rigorously evaluate the effectiveness of {\MBP} across both foundational and large-scale industrial settings. We begin with comprehensive offline analyses to verify that {\MBP} accurately estimates contextual distributions. Building on this, we present results from three challenging real-world scenarios, demonstrating how {\MBP}’s principled debiasing approach leads to measurable improvements in user experience and ecosystem health.



\subsection{Offline Evaluation} 
In the offline evaluation, we begin by assessing the statistical fidelity of the learned contextual distribution, $\mu$ and $\sigma^2$. We then demonstrate that leveraging these contextual distributions enables the proposed relative preference score (RPS) to effectively mitigate biases present in common ranking signals.

\begin{table}[t]
\centering
\caption{Contextual distribution estimation quality measured across accuracy (bias), distributional fit (NLL), and alignment (correlation).}
\label{tab:signal_quality}
\begin{NiceTabular}{l|cc|cc|cc}
\toprule
\multirow{2}{*}{\textbf{Task}} 
& \multicolumn{2}{c|}{\textbf{Bias}} 
& \multicolumn{2}{c|}{\textbf{NLL}} 
& \multicolumn{2}{c}{\textbf{Correlation ($\rho$)}} \\
& ($p, y$) & ($\mu_{\text{{\MBP}}}, p$) 
& $\sigma_{cluster}$ & $\sigma_{\text{\text{\MBP}}}$
& ($p, \mu_{\text{\text{\MBP}}}$) & ($\sigma_{\text{\text{\MBP}}}, \sigma_{\text{cluster}}$) \\ 
\midrule
\textbf{Watch Time} & 0.0014 & -0.0060 & 0.989 & 0.442 & 0.815 & 0.690 \\
\textbf{Plike (Logit)} & 0.0018 & -0.0026 & 5.863 & 1.752 & 0.863 & 0.625 \\ 
\bottomrule
\end{NiceTabular}
\end{table}

\subsubsection{Contextual Distribution Estimation Quality}
We evaluate the quality of {\MBP}'s distribution estimation across three dimensions: accuracy, distributional fit, and alignment in Table \ref{tab:signal_quality}.

\textbf{a. Accuracy (Bias)}: We present both the ranking model bias ($y - p(\mathbf{x})$) and {\MBP} module bias $(p(\mathbf{x}) - \mu)$ as a measure of systematic drift. In Table \ref{tab:signal_quality}, we find that the {\MBP} bias is close to zero (-0.006 \& -0.0026) both in regression task and binary classification task, which confirms $\mu$ acts as an unbiased estimator of the ranking prediction $p$, compared with the baseline ranking bias (0.0014 \& 0.0018) separately.

\textbf{b. Distributional Fit}: 
We employ the negative log-likelihood (NLL) \citep{nix1994estimating, du2021exploration} to measure the goodness-of-fit between the ground truth data and the learned distribution. A lower NLL indicates that $\mu$ and $\sigma$ better capture the underlying uncertainty. As a baseline for variance estimation, we construct $\sigma_{cluster}$ by computing the empirical variance within coarse clusters of ground truth data (e.g., the variance of watch time within different length buckets). Based on the offline analysis in Figure \ref{fig:watch_time_mean_variance_prediction_by_bucket} for watch time and like task, we assume a Gaussian distribution to compute the NLL as follows: 
$$NLL = -\log \mathcal{N}(p | \mu, \sigma^2) = \frac{(p - \mu)^{2}}{2\sigma^{2}} + \frac{1}{2}\log\sigma^2$$
As shown in Table \ref{tab:signal_quality}, {\MBP} yields the lowest NLL, reducing watch time's NLL by more than $50\%$ compared to cluster baselines  (0.442 vs. 0.989). The improvement in like task (1.752 vs. 5.863) also validates the {\MBP}'s ability to model complex heteroscedasticity.

\textbf{c. Alignment}: We utilize the Pearson correlation coefficient ($\rho$) to assess the alignment between {\MBP}’s estimates and the distributional trends observed in the ground truth data:
\begin{itemize}
    \item \textbf{Trend Alignment ($\rho(p, \mu_{\text{\text{\MBP}}})$)}: This metric evaluates the consistency of relative trends, verifying that $\mu$ accurately captures the directional shifts of the ranking model. In both the like and watch time tasks, a correlation greater than $0.81$ demonstrates strong alignment.
    \item \textbf{Uncertainty Alignment ($\rho(\sigma_{\text{\text{\MBP}}}, \sigma_{\text{cluster}})$)}: This metric measures the agreement between the learned variance and the empirical trend of uncertainty in the ground truth data. A high correlation indicates that, for data clusters exhibiting high empirical uncertainty, the learned variance is also elevated. The observed variance correlations of $0.690$ and $0.625$ with the empirical baseline ($\sigma_{\text{cluster}}$) confirm that {\MBP} effectively identifies uncertainty in segments with intrinsically high variance.
\end{itemize}




\begin{table}[h]
\centering
\caption{Correlation analysis between signals and video duration.}
\label{tab:debias_correlation}
\begin{NiceTabular}{l|c||l|c}
\toprule
\multicolumn{2}{c||}{\textbf{Task: Watch Time (Positive Bias)}} & \multicolumn{2}{c}{\textbf{Task: Loop Rate (Negative Bias)}} \\ 
\midrule
\textbf{Method} & \textbf{Correlation ($\rho$)} & \textbf{Method} & \textbf{Correlation ($\rho$)} \\ 
\midrule
$y$ & 0.350 & $y$ & -0.13 \\
Log($y$) & 0.222 &  &  \\
$p(\mathbf{x})$ & 0.514 & $p(\mathbf{x})$ & -0.13 \\
VVP95 & -0.043 & & \\
VVP95 $\times$ NormTS & -0.228 & & \\
\textbf{RPS} & \textbf{0.003} & \textbf{RPS} & \textbf{-0.04} \\ 
\bottomrule
\end{NiceTabular}%
\end{table}
\subsubsection{Bias Mitigation}
A core objective of {\MBP} is to mitigate the correlation between the ranking score and bias attributes (e.g., duration). To evaluate the effectiveness of {\MBP}, we compute the correlation coefficient between the relative preference score (RPS) and duration. A correlation near zero indicates successful debiasing, as the signal becomes independent of the bias attribute. We conduct this analysis using two widely used signals—watch time (inherently positively correlated with duration) and loop rate (inherently negatively correlated)—as shown in Table~\ref{tab:debias_correlation}.

\textbf{Mitigating Positive Duration Bias in Watch Time}.
The watch time label exhibits a strong positive correlation with duration ($\rho=0.350$). Applying a log transformation to the label ($\log(y)$) reduces this correlation to $0.222$, but notable bias remains. When using $\log(y)$ as the prediction target, the ranking model's prediction $p(\mathbf{x})$ actually increases the correlation to $0.514$, indicating that the model amplifies duration bias.

We also report two heuristic production baselines: VVP95 ($\rho=-0.043$) and VVP95 $\times$ NTS ($\rho=-0.228$)\footnote{
1. Video View Percentile (Naive Bucketing): Statistical baselines that bucket videos by duration and calculate the 90th or 95th percentile threshold offline (VVP90 or VVP95). Scores are normalized by checking if a user's watch time exceeds these static thresholds.\\
2. NTS $\times$ VVP95: The normalized time spent (NTS) compensates for user-side bias with a normalized weight. NTS = $\text{sigmoid}(c \times (\text{PredTS} \times (1-\text{pskip}) - \text{Content 7 days Avg TS}))$, where $c$ is a hyperparameter, $\text{pskip}$ is the model-predicted personalized skip rate, and \textit{content 7 days Avg TS} is the content's 7-day average watch time.
}. However, these negative correlations suggest over-correction, where the system inadvertently penalizes longer videos, resulting in an ``inverse'' bias. In contrast, {\MBP} maintains stability ($\rho=0.003$), avoiding both under- and over-correction.

\textbf{Mitigating Negative Duration Bias in Loop Rate}.
The loop rate label is inherently biased against longer videos ($\rho=-0.13$), and the ranking model's prediction $p(\mathbf{x})$ does not alleviate this bias ($\rho = -0.13$). {\MBP} effectively mitigates this issue, reducing the negative correlation to $-0.04$.

These results demonstrate that {\MBP} generalizes across different bias directions and effectively mitigates both positive and negative duration biases.



\subsection{Applications}
\label{sec:application}
In the previous offline experiments, we demonstrated that {\MBP} accurately learns contextual distributions and enables the construction of ranking scores that effectively mitigate inherent biases. In the following case studies, we further show that these bias mitigation capabilities translate into tangible improvements in real-world user experience and overall product health.

\begin{table*}[h]
\centering
\caption{Online A/B testing results across case studies. Note: A boldface means a statistically significant result (p-value< 5\%).}
\label{tab:online_results}
\resizebox{0.9\textwidth}{!}{%
\begin{tabular}{l|l|c|c|c|c|c|c}
\toprule
\multirow{2}{*}{\textbf{Application}} & \multirow{2}{*}{\textbf{Task}} & \multicolumn{6}{c}{\textbf{Metric Lift (\%)}} \\ \cline{3-8} 
 &  & \textbf{Session} & \textbf{View} & \textbf{Watch Time} & \textbf{Share} & \textbf{Like} & \textbf{Breakout} \\ \hline
\multirow{1}{*}{\shortstack[l]{a. Media Length Debias}} 
 & Watch Time & 0.002\% & 0.003\% & \textbf{0.198}\% & 0.44\% & \textbf{0.173}\% & - \\ \hline
\multirow{3}{*}{\shortstack[l]{b. Content  Format Debias}} 
 & Long Time Spent & 0.006\% & -0.011\% & \textbf{0.058}\% & 
-0.071\% & \textbf{-0.119}\% & - \\
 & Like & -0.001\% & \textbf{0.040}\% & 
-0.018\% & 0.035\% & \textbf{0.421}\% & - \\
 & Click & \textbf{0.018}\% & 0.023\% & \textbf{0.034}\% & -0.040\% & 0.007\% & - \\ \hline
\multirow{1}{*}{\shortstack[l]{c. Content Cold Start Debias}} 
 & Watch Time & \textbf{0.011}\% & \textbf{0.135}\% & 0.072\% & 0.049\% & 0.823\% & \textbf{0.190}\% \\ 
\bottomrule
\end{tabular}%
}
\end{table*}

\begin{table*}[t]
\centering
\caption{Engagement efficiency analysis by video length buckets. The table illustrates the traffic redistribution driven by the {\MBP}  model. \textbf{Efficiency Ratio} ($\frac{\%\Delta \text{WT}}{\%\Delta \text{VV}}$) quantifies the quality of the shift: a ratio $<100\%$ during pruning ($\searrow$) indicates the removal of low-value views, while a ratio $>100\%$ during promotion ($\nearrow$) indicates the surfacing of high-retention content.}
\label{tab:length_efficiency}
\resizebox{\textwidth}{!}{%
\begin{tabular}{l|cccc|cccc|ccc|cc}
\toprule
\textbf{Video Length} & \multicolumn{4}{c|}{\textbf{Short-Form (0s - 30s)}} & \multicolumn{4}{c|}{\textbf{Mid-Length (30s - 3m)}} & \multicolumn{3}{c|}{\textbf{Long-Form (3m - 30m)}} & \multicolumn{2}{c}{\textbf{Ultra-Long (30m+)}} \\ 
\midrule
\textbf{Bucket} & 0$\sim$5s & 5$\sim$10s & 10$\sim$15s & 15$\sim$30s & 30$\sim$45s & 45$\sim$60s & 60$\sim$90s & 90$\sim$180s & 3$\sim$5m & 5$\sim$10m & 10$\sim$30m & 30$\sim$60m & 60m+ \\ 
\midrule
\textbf{Watch Time Shift} & -0.64\% & -0.75\% & -0.62\% & -0.19\% & 0.43\% & 0.60\% & 0.77\% & 0.85\% & 1.05\% & 0.73\% & 0.31\% & 0.53\% & -0.65\% \\
\textbf{Video Views Shift} & -0.83\% & -0.80\% & -0.77\% & -0.40\% & 0.12\% & 0.27\% & 0.39\% & 0.41\% & 0.55\% & 0.13\% & -0.23\% & 0.25\% & -0.46\% \\ \hline
\textbf{Efficiency Ratio} & \textbf{77\%} & \textbf{94\%} & \textbf{81\%} & \textbf{47\%} & \textbf{350\%} & \textbf{222\%} & \textbf{198\%} & \textbf{209\%} & \textbf{191\%} & \textbf{562\%} & \textbf{-135\%} & \textbf{>200\%} & \textbf{143\%} \\ \hline
\textbf{Traffic Trend} & $\searrow$ & $\searrow$ & $\searrow$ & $\searrow$ & $\nearrow$ & $\nearrow$ & $\nearrow$ & $\nearrow$ & $\nearrow$ & $\nearrow$ & $\searrow$ & $\nearrow$ & $\searrow$ \\ 
\textbf{Impact} & Pruned & Pruned & Pruned & Pruned & Promoted & Promoted & Promoted & Promoted & Promoted & Promoted & Optimized & Promoted & Pruned \\
\bottomrule
\end{tabular}%
}
\end{table*}

\begin{figure}[ht]
    \centering
    \includegraphics[width=0.9\linewidth]{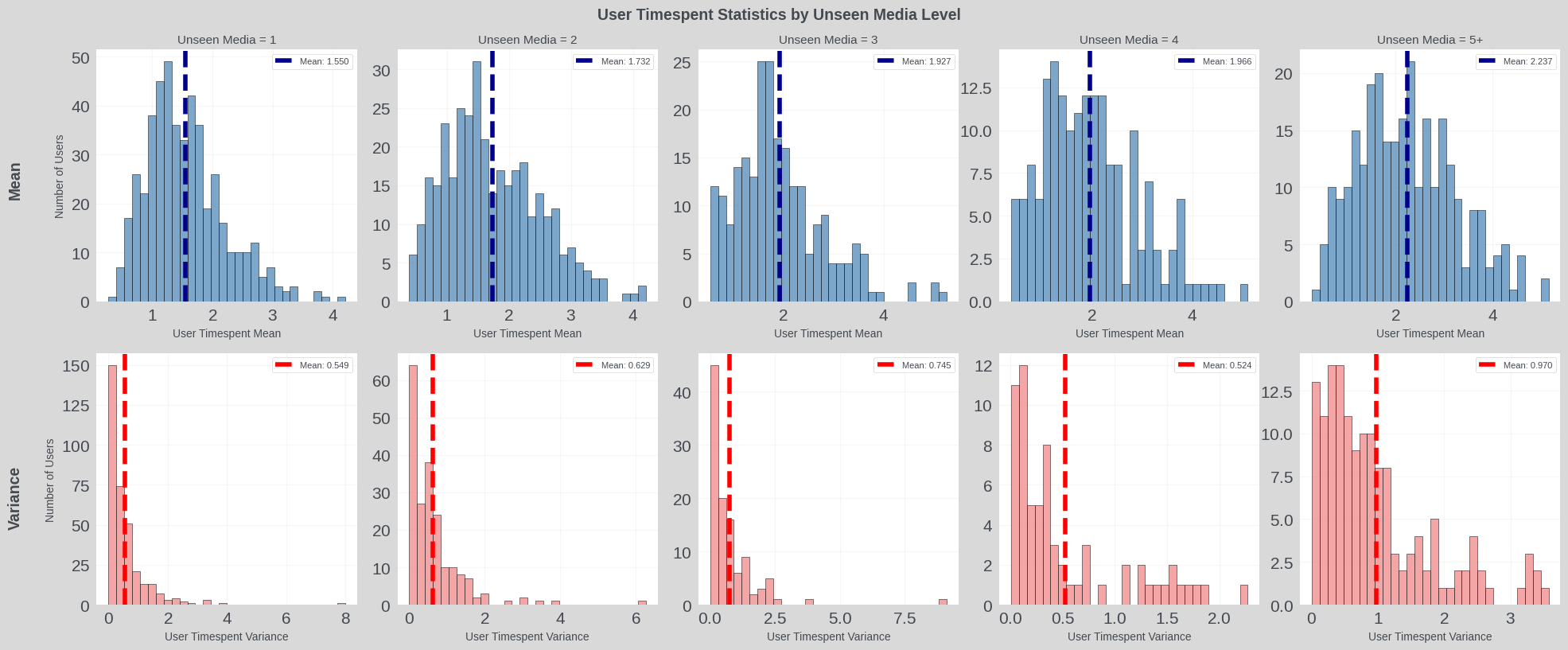}
    \caption{User-level transformed time spent mean and variance distribution stratified by stories length.}
    \label{fig:IG Stories: User-level time spent mean and variance distribution stratified by stories length}
\end{figure}

\subsubsection{Case 1 - Media Length Debias}

In this product application, the ranking system needs to address a persistent bias that disadvantaged longer, multi-media stories, thereby limiting fair competition among diverse content types. Analysis of the contextual mean and variance of transformed watch time, stratified by media length (see Figure~\ref{fig:IG Stories: User-level time spent mean and variance distribution stratified by stories length}), reveals a strong positive correlation: as story length increases, both the expected watch time and the predictive uncertainty also increase substantially.

\begin{itemize}

    \item \textbf{Bias Correction \& Impression Shift:} Previously, time-spent labels systematically penalized multi-media stories. With the introduction of {\MBP}, this bias was corrected, leading to a notable shift in the ecosystem: impressions decreased for single-media stories ($-0.181\%$) and increased for multi-media stories ($+0.58\%$).

    \item \textbf{Removal of Heuristic Constraints:} The debiasing correction resulted in simultaneous gains in both consumption ($+0.198\%$ watch time) and interaction ($+0.173\%$ likes). This concurrent lift demonstrates that, unlike video ranking, story consumption does not inherently involve a trade-off; instead, fair ranking enhances overall relevance by surfacing richer, multi-media content that was previously suppressed by static heuristics.

\end{itemize}

\subsubsection{Case 2 - Content Format Debias}

On certain product surfaces, it is essential for the ranking system to achieve a \textit{uniquely diverse mix} of media types (e.g., photos, videos of varying durations) and to balance content sources (Connected vs. Recommended). Ranking signals must overcome inherent imbalances across formats and account for the diversity in viewer-author relationships.

Beyond previously discussed attributes such as duration, raw predictions often fail to capture the heterogeneous semantics of user interactions—particularly the viewer's relationship with the content author. For example, a ``like'' on a friend's photo may reflect social obligation or personal connection, while a ``like'' on a recommended post typically signals genuine content discovery. Treating these signals uniformly causes downstream ranking to conflate social maintenance with content consumption, potentially misaligning recommendations with users' true discovery preferences.

By directly modeling the $(\text{User} \times \text{Format})$ tuple, {\MBP} enhances multiple tasks to deliver a more balanced content mix and improved user experience:

\begin{itemize}

    \item \textbf{Long Time Spent:} By constructing cohorts based on intersectional features (e.g., $\text{User} \times \text{Type}$), {\MBP} prevents the ranker from over-indexing on video formats due to their intrinsic time-spent advantage over static media like photos, resulting in a significant watch time lift ($+0.058\%$).

    \item \textbf{Incremental Like Value:} By accounting for users' varying tendencies toward different content sources, {\MBP} identifies content with true ``incremental'' like value for each user, leading to a more balanced engagement-to-consumption trade-off.

    \item \textbf{Quality Filtering (Click Task):} By modeling the intersection of format and inventory source (Connected vs. Unconnected), the system identifies ``low confidence'' clicks where $p < \mu - \sigma$. Demoting these effectively filters out clickbait—items with high predicted click probability but low distributional certainty—resulting in a $+0.018\%$ session lift and $+0.034\%$ watch time lift.

\end{itemize}


\subsubsection{Case 3 - Content Cold Start Debias}

In addition to content length and format, a significant source of bias in ranking systems arises from the number of views a piece of content has already accumulated. This often disadvantages newer content—a challenge known as the content cold start problem. In this case study, {\MBP} addresses exposure bias for cold start content by modeling the tuple $(\text{User} \times \text{length} \times \text{views})$, capturing the nuanced interactions between user characteristics, content length, and prior exposure.

{\MBP} delivers strong improvements in key metrics: breakout ($+0.190\%$), session ($+0.011\%$), and views ($+0.135\%$). Cold start items typically suffer from high uncertainty due to limited interaction history. By explicitly modeling this uncertainty through the watch time task, {\MBP} can identify items with high potential (i.e., high predicted mean $\mu$ and high variance $\sigma$ relative to their view count).

    
    


\subsection{Engagement Efficiency Analysis}
To understand how {\MBP} reshapes the content ecosystem, we analyze the traffic distribution shift across video length buckets. We define \textit{engagement efficiency} as the ratio between the relative lift in watch time and video views (VV): $\text{Efficiency} = \%\Delta \text{WT}/\%\Delta \text{VV}$. As detailed in Table \ref{tab:length_efficiency}, the framework optimizes the ecosystem through two distinct mechanisms:
\begin{itemize}
    \item \textbf{Pruning Low-Value Short-Videos (0s--30s):} The model induces a negative shift in short-video. Crucially, the decline in views (e.g., -0.83\% for 0-5s) consistently exceeds the decline in watch time (-0.64\%). This results in an efficiency ratio of $<100\%$ in the negative domain, indicating that the model selectively prunes views — short, looping clips that generate views but minimal user attention.

    \item \textbf{Promoting High-Retention Long-Videos (30s+):} For videos with duration > 30s, the model drives a positive shift with high efficiency. For instance, in the 5-10m bucket, a modest +0.13\% increase in views yields a massive +0.73\% gain in watch time (efficiency ratio = 562\%). This confirms that {\MBP} successfully identifies and surfaces high quality content previously suppressed by duration bias that delivers disproportionately high retention when exposed.
\end{itemize}

\subsection{Special Handling for Binary Signals}
A critical challenge in {\MBP} framework is in accurately modeling the variance of sparse binary events (e.g., shares or comments) where the background click-through rate (CTR) is extremely low ($\approx$1\%). In the probability space, these events are heavily clustered near zero. This creates a ``saturated tail'' phenomenon: the absolute difference between a user with $p=0.001$ and $p=0.01$ like rate is numerically negligible (0.009) compared with the difference between $p=0.2$ and $p=0.4$, yet it represents a 10x difference vs. 2x difference in user preference intensity. Consequently, applying DMoM directly on probabilities results in vanishing gradients and an inability to distinguish signal from noise.

To resolve this, we introduce  \textit{logit-space projection} method. Instead of modeling the bounded probability, we project the prediction into the unbounded latent space $\mathbb{R}$ by estimating the statistics of the logit:
\begin{equation*} 
\text{logit}(p(\mathbf{x})) = \ln\left(\frac{p(\mathbf{x})}{1-p(\mathbf{x})}\right)
\end{equation*}
By setting the target $\text{logit}(p(\mathbf{x}))$ (the raw logit output of the main ranking model), we effectively ``stretch'' the compressed probability interval near zero into a vast, quasi-symmetric manifold, check distribution in Figure \ref{fig:plike_mean_variance_prediction_by_bucket}. This transformation ensures numerical stability and prevents the asymmetric penalty issue, allowing the moments to capture the true relative spread of user preference that would otherwise be obscured by the sigmoid nonlinearity.

\section{Related Work}
\label{sec: related work}

\subsection{Debiasing in Recommender Systems}
Recommender systems rely on interaction signals to model preferences, yet these signals are inherently confounded by biases such as selection, exposure, and position. Without intervention, models risk amplifying feedback loops where popular items suppress niche content regardless of true user preference. 
Traditional mitigation strategies categorize into: (1) Inverse Propensity Weighting, which re-weights training samples by the inverse of their exposure probability \citep{chen2023bias}; (2) Causal Intervention, utilizing backdoor adjustment to block confounding paths \citep{wang2021deconfounded}; and (3) Regularization, such as In-Batch Balancing Regularization (IBBR), which adds fairness penalty terms to loss functions to mitigate ranking disparity among subgroups \citep{li2022debiasing}. However, these methods typically focus on correcting point-wise estimates, often failing to provide the distributional context necessary for calibrating relative preference across heterogeneous content formats.

\subsection{Duration Bias and Watch Time Prediction}
In video recommendation, addressing the ``duration bias'' is critical for fair ranking.
Early work like \citep{covington2016deep} introduced weighted logistic regression to modify training odds, approximating expected watch time by weighting samples with their duration. Recent research has evolved toward causal and adversarial frameworks:
causal adjustment D2Q \citep{zhan2022deconfounding} treats duration as a confounder affecting both exposure and watch time. It utilizes backdoor adjustment and duration-group quantile modeling to remove exposure bias while preserving the natural correlation with watch time. AlignPxtr \citep{lin2025alignpxtr} further advances this by aligning predicted behavior distributions via quantile mapping to ensure independence between bias factors and user interest.
DVR \citep{zheng2022dvr} introduces ``watch time gain'' optimization via adversarial learning to decouple duration from preference. D2Co \citep{zhao2023uncovering} simultaneously corrects for duration bias and ``noisy watching'' (user's dwelling on content they dislike) using a mixture of latent distributions.
SWaT \citep{yang2025swat} adopts a user-centric perspective, modeling watch time through statistical buckets derived from specific user behavior assumptions (e.g., random picking vs. sequential watching) to handle non-stationary viewing probabilities.
While effective at correcting mean estimates, these approaches often lack the granular uncertainty quantification required to distinguish whether a high prediction is a signal of confidence or merely an artifact of high variance in long-form content.
Unlike these methods which focus on correcting specific bias types, {\MBP} provides a unified framework applicable to any bias defined by a partial feature set.

\subsection{Uncertainty and Distributional Learning}
{\MBP} moves beyond point estimation to distributional learning, estimating the statistical properties of the model's behavior to enable relative scoring.
Quantile regression techniques like CQE \citep{lin2024conditional} apply deep quantile regression to estimate the full conditional distribution of watch time. By predicting multiple quantiles (e.g., $P_{95}$), these methods capture the inherent uncertainty and heterogeneity of user engagement, which point estimates (mean) fail to represent. Similarly, non-crossing quantile networks \citep{shen2025deep} enforce logical consistency in distribution estimation.
Recent works also explore conformal prediction \citep{cherian2024large}, Epinet \citep{jeon2024epinet}, and prediction-powered inference (PPI) \citep{angelopoulos2023prediction} to construct rigorous confidence intervals. While these methods focus on bounding prediction uncertainty, {\MBP} leverages distributional statistics to enable relative scoring across heterogeneous cohorts.

\section{Conclusion and Future Work}
In this paper, we propose the {\MBP} framework, a generalized solution for signal debiasing in large-scale recommender systems across user, content, and model dimensions. By shifting from absolute point-wise estimation to relative distributional modeling, {\MBP} constructs an arbitrary feature set to debias existing signals by estimating the contextual mean and contextual variance with maximal efficiency and reliability. This is achieved by adding a lightweight computational module with zero impact on the existing ranking model.
Extensive offline analysis demonstrates the fidelity of the estimated contextual mean and contextual variance and validates the debiasing effect through correlation analysis. In addition, we apply {\MBP} to three challenging real-world scenarios  
over a platform serving billions of users which confirms {\MBP} significantly improving long-term engagement metrics. In the future, the {\MBP} framework can be extended to explore automated bias discovery techniques to dynamically identify bias features.

\bibliographystyle{assets/plainnat}
\bibliography{mbp}

\clearpage
\newpage
\beginappendix
\section{Estimation}
In this section, we present the another option of estimating the distribution through quantile estimation.

\subsection{Distribution Estimation}
\label{sec:distribution estimation}
Quantile regression estimates contextual quantiles $Q_\tau(p(\mathbf{x})|\mathbf{x}')$ for a specified quantile level $\tau \in (0, 1)$, providing a distribution-free approach to uncertainty quantification. The pinball loss function imposes asymmetric penalties based on the direction of prediction errors:
\begin{equation}
    \min_{q_{\tau}}L_{\tau}(q_{\tau}) = \left[\tau - \mathbb{I}(p(\mathbf{x}) < q_{\tau})\right](p(\mathbf{x}) - q_{\tau})
\end{equation}
where $\mathbb{I}(\cdot)$ is the indicator function. Note that when $\tau = 0.5$, this reduces to $L_1$ loss (median regression).

Quantile regression is a distribution-free method that is robust to outliers due to its reliance on contextual quantiles, when the distribution is not unimode. It directly provides interpretable percentile-based uncertainty estimates suitable for top-$k$\% selection in recommender system.
However, estimation accuracy deteriorates for extreme quantiles (e.g., $\tau = 0.01$ or $\tau = 0.99$) due to data sparsity in the distribution tails, which are important to the system. The quantile level $\tau$ must be specified during training and remains fixed post-deployment, limiting adaptability when operational requirements shift (e.g., changing from $\tau$ to $\tau'$ requires retraining). In addition, the quantile method can be an optional addition to the existing {\MBP} framework.

\section{Additional Experiment}
\subsection{Video Distributional Fitting Analysis}
To empirically validate the effectiveness of the contextual baseline estimation, we present the alignment between the {\MBP}-estimated statistics ($\mu_{\text{\MBP}}, \sigma^2_{\text{\MBP}}$) and the empirical statistics of the main model's predictions $p(\mathbf{x})$ across video duration buckets.

\begin{figure*}[t]
    \centering
    \begin{minipage}{0.48\linewidth}
        \centering
        \includegraphics[width=\linewidth]{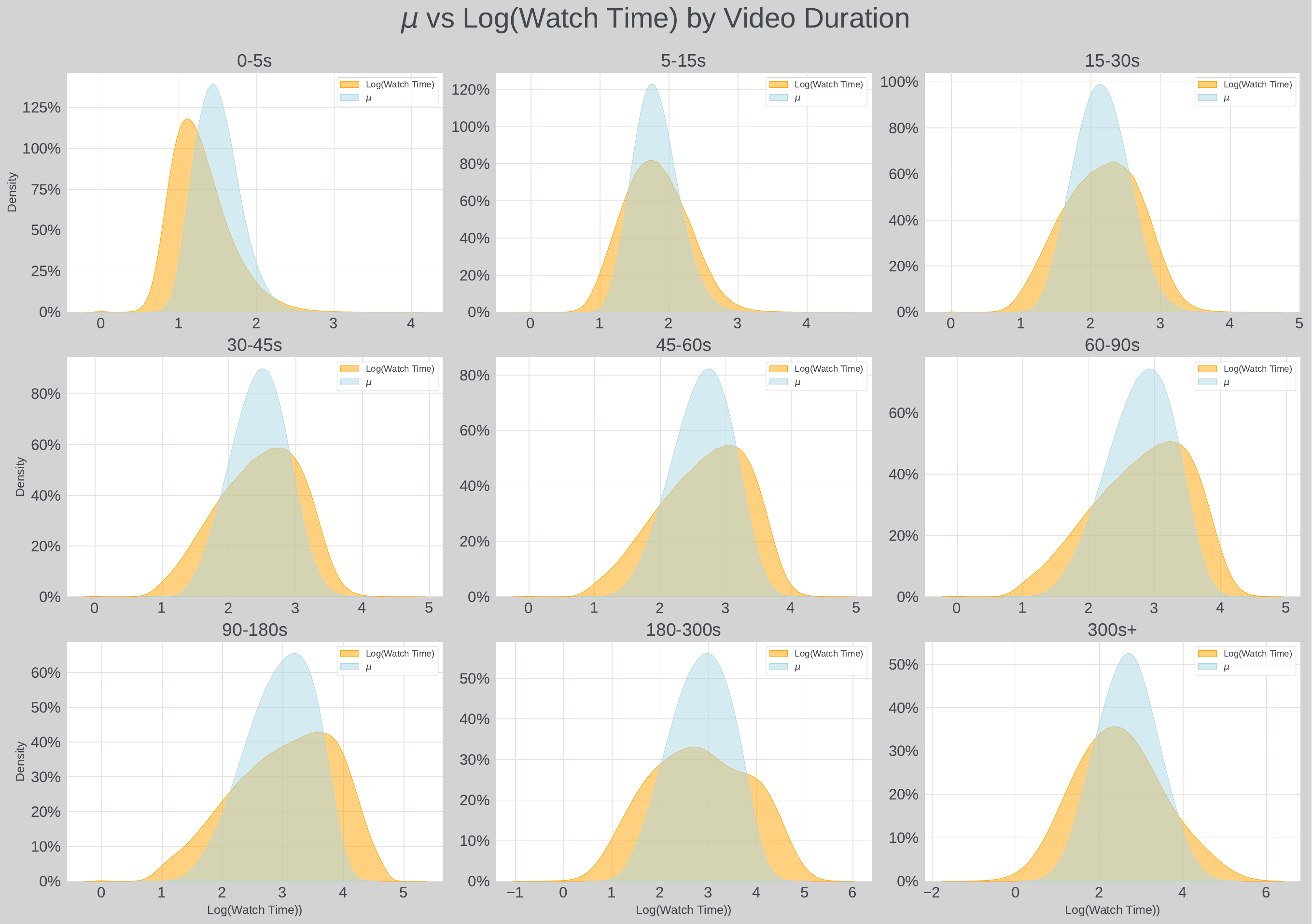}
        \centerline{(a) $p(\mathbf{x})$ vs $\mu_{\text{\MBP}}$}
    \end{minipage}
    \hfill
    \begin{minipage}{0.48\linewidth}
        \centering
        \includegraphics[width=\linewidth]{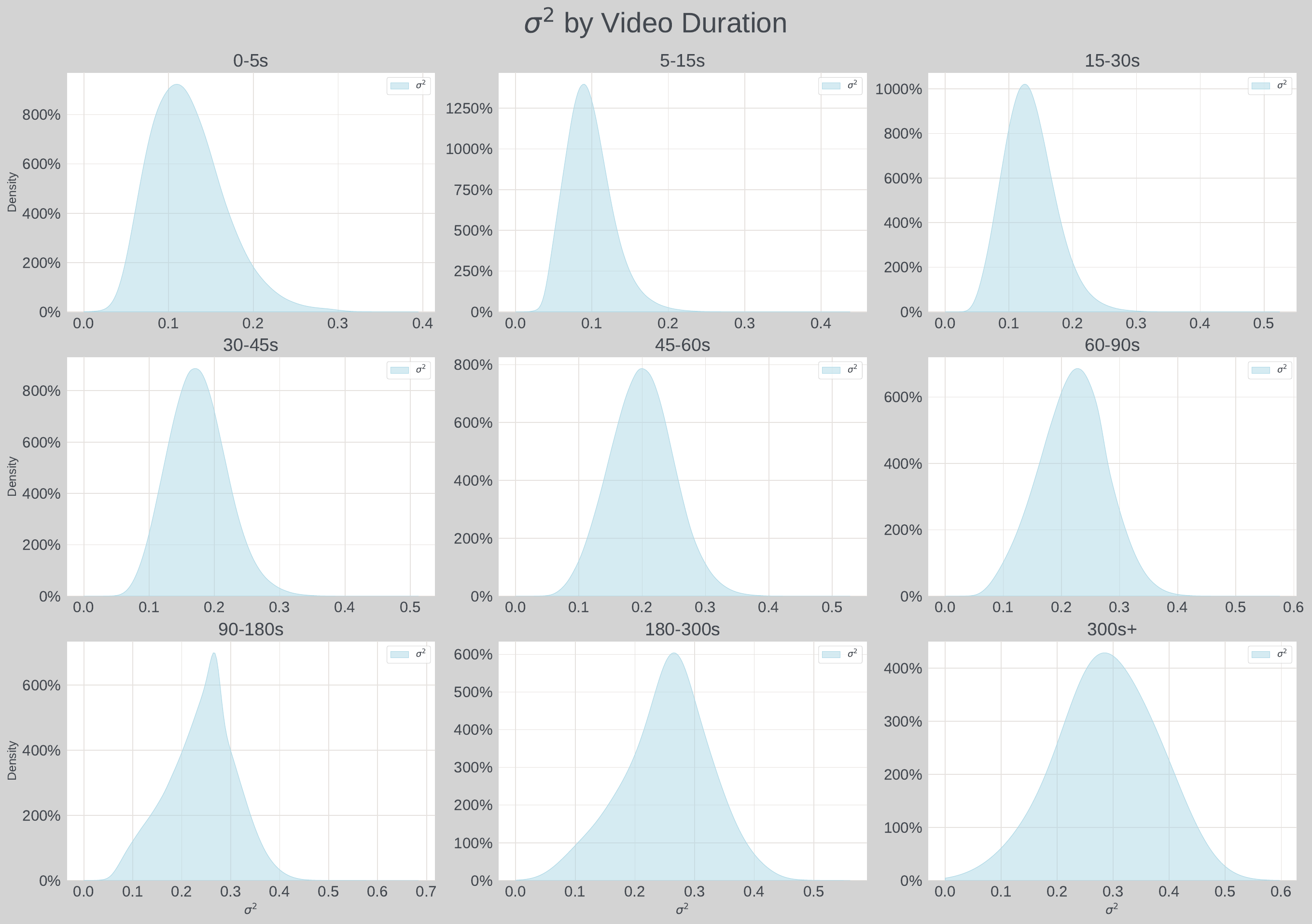}
        \centerline{(b) $\sigma_{\text{\MBP}}^2$}
    \end{minipage} 
    \caption{Analysis of {\MBP} predictions across duration buckets: (a) The predicted mean $\mu_{\text{\MBP}}$ closely tracks the empirical distribution $p(\mathbf{x})$; (b) The predicted variance $\sigma_{\text{\MBP}}^2$ effectively captures the uncertainty within each bucket.}
    \label{fig:watch_time_mean_variance_prediction_by_bucket}
\end{figure*}

\begin{figure*}[t]
    \centering
    \begin{minipage}{0.48\linewidth}
        \centering
        \includegraphics[width=\linewidth]{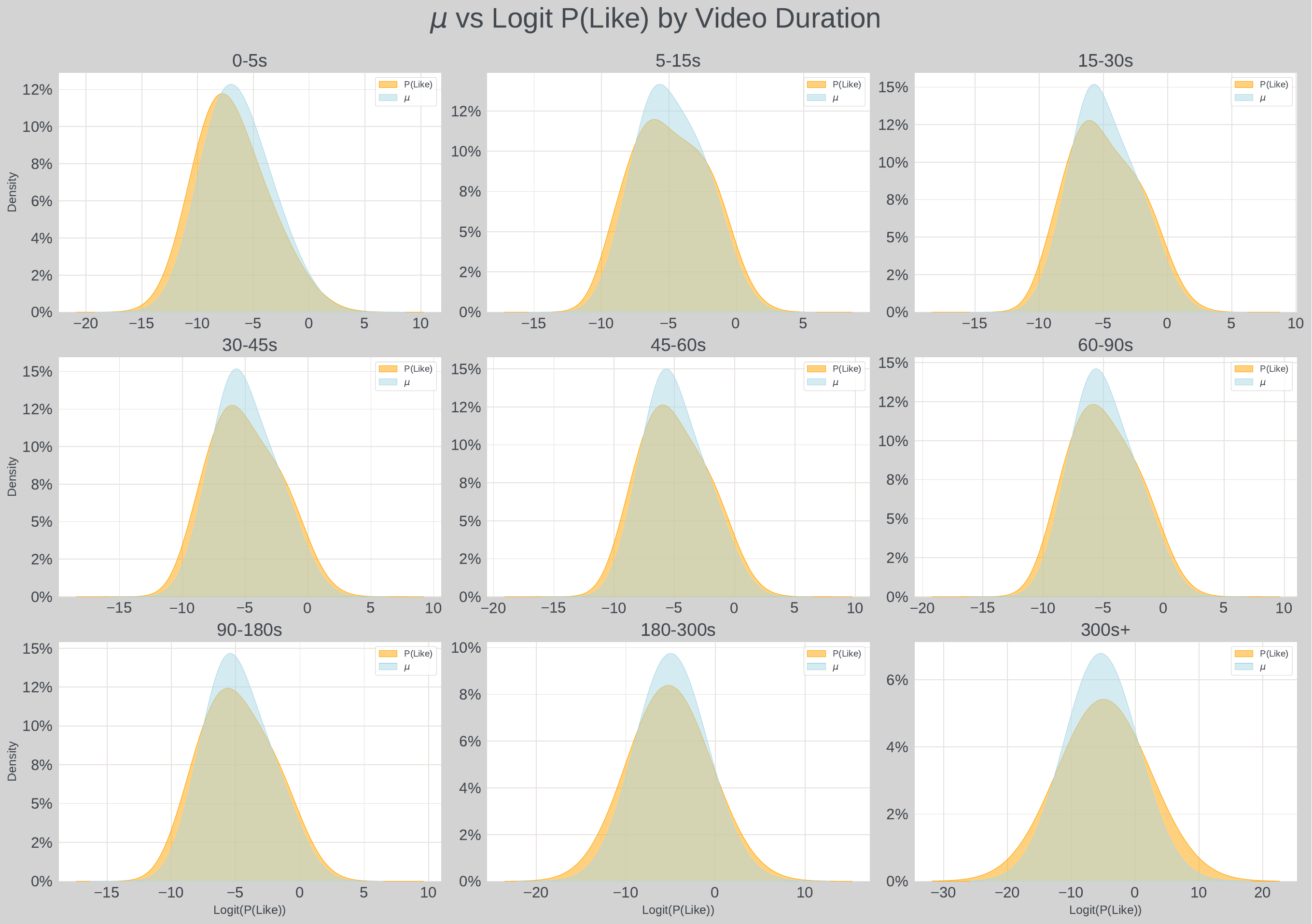}
        \centerline{(a) $p(\mathbf{x})$ vs $\mu_{\text{\MBP}}$ (Logit Space)}
    \end{minipage}
    \hfill
    \begin{minipage}{0.48\linewidth}
        \centering
        \includegraphics[width=\linewidth]{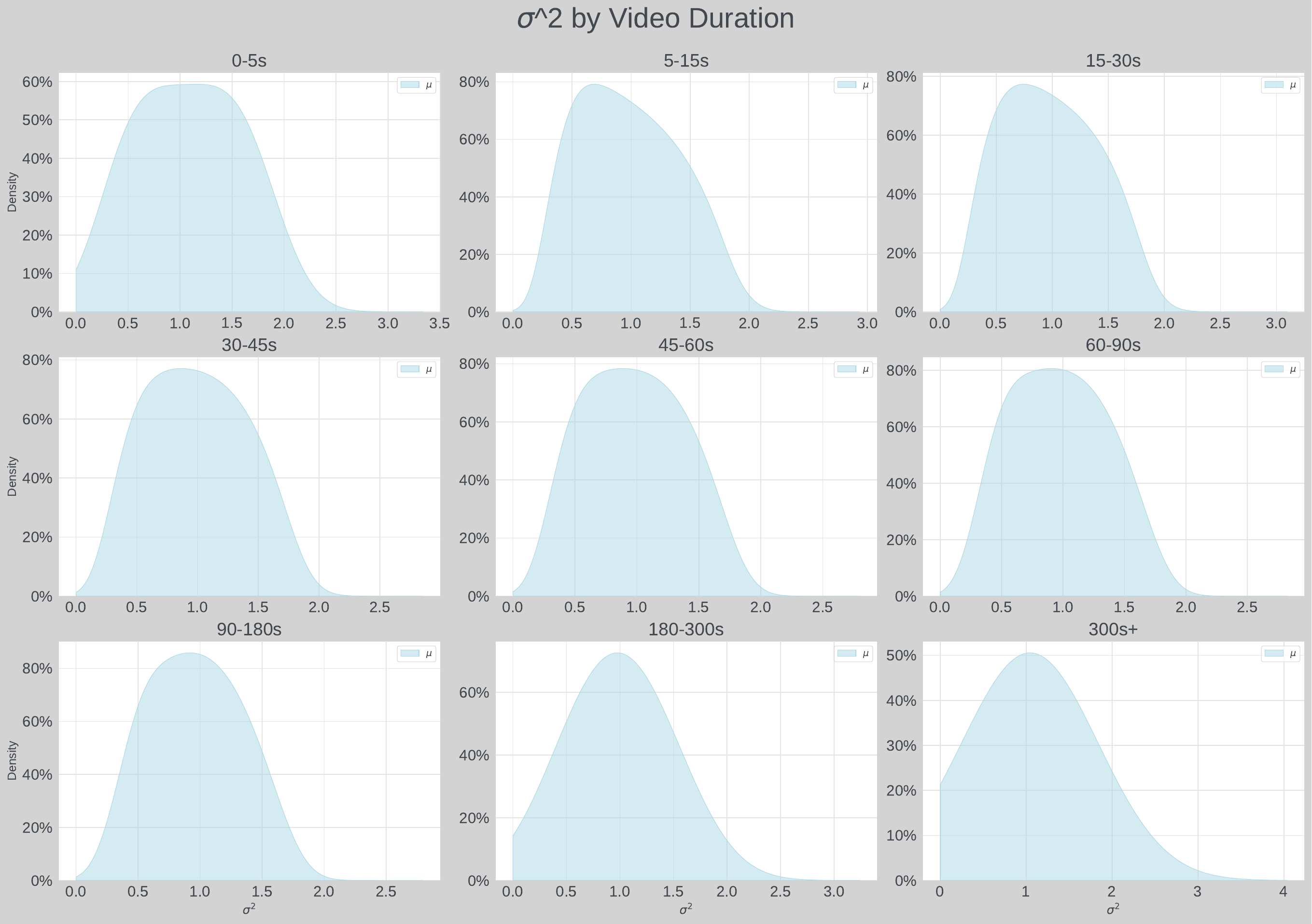}
        \centerline{(b) $\sigma_{\text{\MBP}}^2$ (Logit Space)}
    \end{minipage}
    \caption{Analysis of {\MBP} predictions for Likes in logit space: (a) The predicted mean $\mu_{\text{\MBP}}$ accurately tracks the main model's logit prediction $p(\mathbf{x})$; (b) The predicted variance $\sigma_{\text{\MBP}}^2$ quantifies the uncertainty of user preference intensity.}
    \label{fig:plike_mean_variance_prediction_by_bucket}
\end{figure*}

\textbf{Mean Estimation.} As illustrated in Figure \ref{fig:watch_time_mean_variance_prediction_by_bucket}(a) for Watch Time and Figure \ref{fig:plike_mean_variance_prediction_by_bucket}(a) for Like Probability (in Logit Space), the {\MBP}-predicted mean $\mu_{\text{\MBP}}$ exhibits near-perfect alignment with the empirical average of the main model's predictions. This confirms that the {\MBP} module successfully ``profiles'' the bias curve—accurately capturing the natural logarithmic increase in predicted watch time as duration grows—without requiring ground-truth labels for the auxiliary branch.

\textbf{Variance Estimation.} Figures \ref{fig:watch_time_mean_variance_prediction_by_bucket}(b) and \ref{fig:plike_mean_variance_prediction_by_bucket}(b) demonstrate the variance estimation capabilities. The {\MBP}-predicted variance $\sigma^2_{\text{\MBP}}$ closely tracks the empirical variance of $p(\mathbf{x})$. Crucially, for sparse binary targets like Likes, fitting in the logit space allows the model to capture significant heteroscedasticity in user preference intensity that is typically compressed by the sigmoid function. This accurate variance estimation is vital for the normalization step in RPS construction, ensuring relative scores are properly calibrated against the intrinsic uncertainty of each duration cohort.


\end{document}